\documentclass[runningheads]{llncs}

\usepackage[mobile]{eccv}

\usepackage{eccvabbrv}

\usepackage{graphicx}
\usepackage{booktabs}
\usepackage{multirow}

\usepackage[accsupp]{axessibility}  % Improves PDF readability for those with disabilities.

\usepackage[pagebackref,breaklinks,colorlinks]{hyperref}
\usepackage{orcidlink}

\usepackage{colortbl}
\usepackage{array}
\usepackage{tabularx}
\usepackage{wrapfig}

\definecolor{top1}{RGB}{255,179,179}
\definecolor{top2}{RGB}{255,217,179}
\definecolor{top3}{RGB}{255,255,179}

\newcommand{\camready}[1]{#1}

\newcommand\blfootnote[1]{%
  \begingroup
  \renewcommand\thefootnote{}\footnote{#1}%
  \addtocounter{footnote}{-1}%
  \endgroup
}

\newenvironment{tightcenter}{%
  \setlength\topsep{0pt}
  \setlength\parskip{0pt}
  \begin{center}
}{%
  \end{center}
}

\begin{document}

\title{CoherentGS: Sparse Novel View Synthesis with Coherent 3D Gaussians} 

\titlerunning{CoherentGS}

\author{Avinash Paliwal\inst{1,2*} \and
Wei Ye\inst{2} \and
Jinhui Xiong\inst{2} \and
Dmytro Kotovenko\inst{3} \and
Rakesh Ranjan\inst{2} \and
Vikas Chandra\inst{2} \and
Nima Khademi Kalantari\inst{1}}

\authorrunning{A.~Paliwal et al.}

% Springer Heidelberg, Tiergartenstr.~17, 69121 Heidelberg, Germany

\institute{Texas A\&M University \and
% \email{\{avinashpaliwal, nimak\}@tamu.edu}\and
Meta Reality Labs \and LMU Munich\\
\vspace{0.8em}
\url{https://people.engr.tamu.edu/nimak/Papers/CoherentGS}
\vspace{-1.8em}}

\maketitle

\begin{abstract}
The field of 3D reconstruction from images has rapidly evolved in the past few years, first with the introduction of Neural Radiance Field (NeRF) and more recently with 3D Gaussian Splatting (3DGS). The latter provides a significant edge over NeRF in terms of the training and inference speed, as well as the reconstruction quality. Although 3DGS works well for dense input images, the unstructured point-cloud like representation quickly overfits to the more challenging setup of extremely sparse input images (e.g., 3 images), creating a representation that appears as a jumble of needles from novel views. To address this issue, we propose regularized optimization and depth-based initialization. Our key idea is to introduce a structured Gaussian representation that can be controlled in 2D image space. We then constraint the Gaussians, in particular their position, and prevent them from moving independently during optimization. Specifically, we introduce single and multiview constraints through an implicit convolutional decoder and a total variation loss, respectively. With the coherency introduced to the Gaussians, we further constrain the optimization through a flow-based loss function. To support our regularized optimization, we propose an approach to initialize the Gaussians using monocular depth estimates at each input view. We demonstrate significant improvements compared to the state-of-the-art sparse-view NeRF-based approaches on a variety of scenes.
\keywords{Sparse View Synthesis \and 3D Gaussian Splatting \and Implicit Decoder}
\end{abstract}

\blfootnote{*The work was primarily done during an internship at Meta Reality Labs.}

\section{Introduction}
\label{sec:intro}
\begin{figure}[t]
    \centering
  \includegraphics[width=0.8\linewidth]{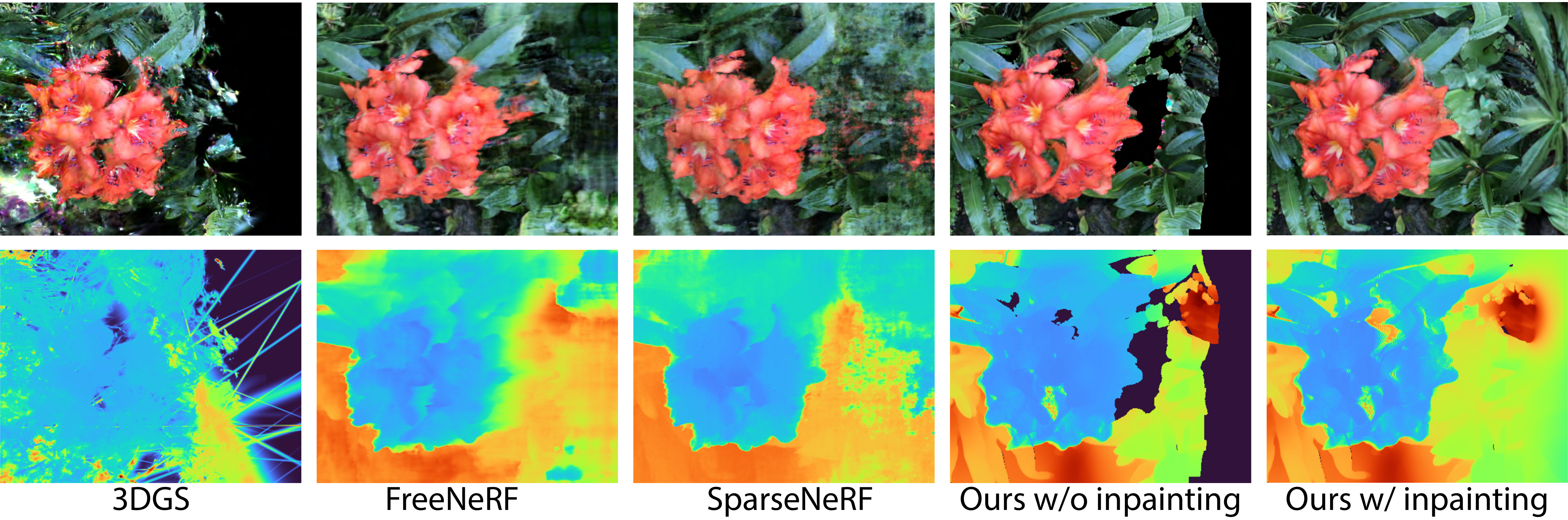}
  % \vspace{-0.2in}
    \caption{For sparse input views, the quality of 3DGS deteriorates. Notable artifacts are observed in the results of the NeRF-based methods by Yang et al.~\cite{yang2023freenerf} (FreeNeRF) and Wang et al.~\cite{wang2023sparsenerf} (SparseNeRF). Our approach (``Ours w/o inpainting'') yields high-quality synthesized views. Note that our constraints do not allow the Gaussians to move freely in the 3D space. As a result, our approach does not reconstruct the areas that are occluded in all the input images. This is an advantage of our technique over other methods that fill in these areas with blurry and repetitive structure, as we can identify and inpaint these regions and produce realistic hallucinated details. As a proof of concept, we inpaint these regions using a diffusion model and project them to 3D using monocular depth. As shown on the right, the hallucinated details and their corresponding depth are reasonable.}
    \label{fig:3dgs_comparison}
    % \vspace{-0.2in}
\end{figure}

In recent years, the field of 3D reconstruction from posed images has been in the spotlight with the integration of learned scene representations, such as Multi-Plane Images (MPI)~\cite{zhou2018stereo} and Neural Radiance Field (NeRF)~\cite{mildenhall2020nerf}. More recently, an explicit representation known as 3D Gaussian Splatting (3DGS)~\cite{kerbl3Dgaussians} has been introduced that significantly improves the training speed while providing real-time inference and better 3D reconstruction quality. However, it struggles to generate a good representation given a sparse set of training images. In such cases, the representation severely overfits to the training views and appears as a collection of semi-random anisotropic blobs from novel views, as shown in Fig.~\ref{fig:3dgs_comparison}.

While utilizing 3DGS for sparse input view synthesis is under-explored, several NeRF-based methods have been proposed to tackle this task. Since reconstruction from sparse images is an ill-posed problem, these approaches often employ various regularizations to constrain the optimization~\cite{Niemeyer2021Regnerf, yang2023freenerf, wang2022sparsenerf, seo2023flipnerf, SN2023ViPNeRFVP}. However, as shown in Fig.~\ref{fig:3dgs_comparison}, these state-of-the-art NeRF-based approaches produce sub-optimal results as their regularizations do not provide sufficient constraints for a reasonable 3D reconstruction. Additionally, most of these approaches rely on the coherency of the implicit representation learned by a neural network and are not directly applicable to 3DGS with an explicit unstructured representation.

Our key idea is to augment the explicit unstructured representation with coherency, by constraining the movement of the Gaussian blobs during optimization. Because of the unstructured nature of the Gaussians, constraining them in the 3D space is difficult. To overcome this issue, we propose to assign a single Gaussian to every pixel of each input image and enforce single and multiview constraints in 2D image space. Specifically, we force the Gaussians of each image with similar depth to move coherently using an implicit decoder~\cite{bemana2020xfields, paliwal2023implicit, liu2018an}. To enforce constraints over Gaussians across different views, we ensure the rendered depth using all the Gaussians are smooth through a total variation loss. With the introduced coherency, we additionally propose a flow-based loss to ensure the position of the Gaussians of the corresponding pixels in two images are similar.

Furthermore, to help with the optimization, we propose to initialize the position of Gaussians using an existing monocular depth prediction model. While these models provide high-quality depth estimates, single image depth is relative and lacks consistency across views. Our depth-based initialization properly positions the Gaussians in the world space, while our regularized optimization encourages the updates, particularly to the positions, to be coherent and smooth. This allows us to reconstruct high-quality texture and geometry, as shown in Fig.~\ref{fig:3dgs_comparison}. As an added benefit of preventing the Gaussians to freely move in the 3D space, we can easily identify and inpaint the occluded regions to produce high-quality hallucinated texture and geometry (see Fig.~\ref{fig:3dgs_comparison} - right).

In summary, we make the following key contributions:
\begin{itemize}
  \setlength{\itemindent}{1.2em}
  \item We present an approach to perform 3D reconstruction using 3DGS from extremely sparse set of inputs.
  \item We propose a structured Gaussian representation and introduce coherency using various regularizations. 
  \item We introduce a depth-based initialization of 3D Gaussians that complements the regularized optimization.
\end{itemize}
    
\section{Related Work}
\label{sec:relatedwork}
In this section, we mainly focus on the closely related work by discussing the radiance field approaches and sparse novel view synthesis methods that utilize this concept.

\subsection{Radiance Fields}
The introduction of Neural Radiance Field (NeRF)~\cite{mildenhall2020nerf}, an optimization based approach that leverages implicit neural networks, has revolutionized scene reconstruction and novel view synthesis. A large number of techniques have presented various ways to improve the rendering quality~\cite{barron2021mipnerf, barron2022mipnerf360, barron2023zipnerf, guo2022nerfren, attal2022learning, suhail2022light, verbin2022refnerf, wang2023f2nerf}, generalization~\cite{chen2021mvsnerf, chibane2021stereo, johari2022geonerf, varma2023is, wang2021ibrnet, yu2020pixelnerf}, performance on dynamic scenes~\cite{du2021neural, bemana2020xfields, paliwal2023implicit, li2020neural, pumarola2020dnerf, tretschk2021nonrigid, gao2021DynNeRF}, and the need for precise camera poses~\cite{yen2020inerf, wang2021nerfmm, xia2022sinerf, lin2021barf, chng2022gaussian, bian2022nopenerf}. Moreover, the success of this approach has also led to its use as the representation of choice in 3D generative modeling~\cite{cao2023dreamavatar, chan2021pigan, gu2022stylenerf, hoellein2023text2room, poole2022dreamfusion}. 

However, NeRF models based on MultiLayer Perceptron (MLP) utilize the neural network to implicitly encode the scene density and view-dependent color. To render a pixel color, the MLP is queried several times along the sampled ray which leads to slow optimization (hours) and rendering (seconds per frame). While several approaches have been proposed to improve the optimization and inference speed~\cite{yu_and_fridovichkeil2021plenoxels, sun2022direct, mueller2022instant, chen2022tensorf, garbin2021fastnerf, reiser2021kilonerf}, they often achieve the speed up by sacrificing the rendering quality. Recently, Kerbl et al.~\cite{kerbl3Dgaussians} achieved a breakthrough with the 3D Gaussian Splatting (3DGS) approach which enables fast optimization with high-quality real-time rendering. We introduce a novel approach that enables application of 3DGS to the sparse input setting.

\subsection{Sparse Novel View Synthesis}
An interesting and more practical variation of the view synthesis problem is to utilize a sparse set of input images. NeRF and 3DGS require tens of images to render high quality novel views. With only a few images, these methods quickly overfit to the input images and produce unsatisfactory novel view results. Several NeRF-based approaches~\cite{Niemeyer2021Regnerf, yang2023freenerf, wang2022sparsenerf, seo2023flipnerf, SN2023ViPNeRFVP, kangle2021dsnerf, jain2021putting, yu2020pixelnerf} tackle this problem by introducing various regularizations. Specifically, RegNeRF~\cite{Niemeyer2021Regnerf} proposes a geometry and color regularization from unobserved viewpoints. DS-NeRF~\cite{kangle2021dsnerf} relies on sparse 3D points from COLMAP for depth supervision. ViP-NeRF~\cite{SN2023ViPNeRFVP} computes a visibility prior to impose a multi-view constraint during optimization. FreeNeRF~\cite{yang2023freenerf} uses a coarse-to-fine refinement scheme by gradually increasing the positional encoding frequencies. SparseNeRF~\cite{wang2022sparsenerf} introduces a local depth ranking and spatial continuity regularization based on monocular depth. 

While these approaches produce impressive results, their regularization in extremely sparse settings (e.g., 3 images) is not able to provide sufficient constraints. Therefore, as shown in Fig.~\ref{fig:3dgs_comparison}, their results exhibit significant blurring in these scenarios. Additionally, the regularizations in most of these techniques are sparse and rely on the coherency of the MLP to propagate the constraints to the other regions. Unfortunately, these regularizations are not directly applicable to 3DGS which is a discrete and unstructured representation. We propose to introduce coherency in the 3DGS representation by utilizing the combination of an implicit decoder and total variation loss.

Concurrent to our technique, Zhu et al.~\cite{zhu2023fsgs} (FSGS), Xiong et al.~\cite{xiong2023sparsegs} (SparseGS) \camready{and Li et al.~\cite{li2024dngaussian} (DNGaussian)} propose to utilize 3DGS for sparse view synthesis. However, unlike our method, they do not present a way to enforce coherency between nearby Gaussians, producing occasional floaters. Additionally, these methods fill in the regions that are occluded in all the input images with elongated Gaussians producing blurry results. In contrast, our approach allows us to identify and inpaint the occluded regions and produce high-quality hallucinated details (see Fig.~\ref{fig:3dgs_comparison} - right).

\section{Background}
\label{sec:gauss_splat}

In this section, we provide a brief introduction to the 3DGS~\cite{kerbl3Dgaussians} approach. This technique represents a 3D scene using a dense set of anisotropic 3D Gaussians, enabling fast and differentiable rendering via $\alpha$-blending. Each Gaussian is parameterized by a set of attributes such as position $\mathbf{x}$, color $\mathbf{c}$, covariance matrix $\Sigma$~\cite{yifan2019differentiable}, and opacity value $\alpha$. Similar to a typical differentiable point-based approach~\cite{kopanas2022neural}, the image formation model for the volumetric rendering can be written as:
\begin{equation}
	\label{eq:rendered_color}
	R_{\Sigma, \alpha, \mathbf{x}, \mathbf{c}}(\mathbf{p}) = \sum_{i \in \mathcal{N}(\mathbf{p})}
	\mathbf{c}_{i}\gamma_i
	\prod_{j=1}^{i-1}(1-\gamma_j) \;\text{, where}\; \gamma_k=f(\Sigma_k, \alpha_k, \mathbf{x}_k, \mathbf{p}).
\end{equation}
Here, $\mathcal{N}(\mathbf{p})$ is the number of ordered points overlapping the pixel of interest $\mathbf{p}$. Moreover, the function $f(\Sigma, \alpha, \mathbf{x}, \mathbf{p})$ computes the effective opacity by evaluating the Gaussian at $\mathbf{p}$ and multiplying it with the global opacity $\alpha$. Note that in practice, 3DGS decomposes $\Sigma$ into scaling and rotation components, and uses spherical harmonics to represent the view-dependent color. However, we will continue using $\Sigma$ and $\mathbf{c}$ in the following sections for simplicity of notation.

During optimization, optimal Gaussian parameters and colors are obtained by minimizing the following objective:
\begin{equation}
    \label{eq:3dgs_opt}
	\Sigma^*, \alpha^* \mathbf{x}^*, \mathbf{c}^* = \arg \min_{\Sigma, \alpha, \mathbf{x}, \mathbf{c}}\sum_{\mathbf{p} \in \mathcal{P}}\,\mathcal{L}(R_{\Sigma, \alpha, \mathbf{x}, \mathbf{c}}(\mathbf{p}), R(\mathbf{p})),
\end{equation}
where $\mathcal{P}$ contains all pixels of every input image, $R_{\Sigma, \alpha, \mathbf{x}, \mathbf{c}}(\mathbf{p})$ and $R(\mathbf{p})$ correspond to the rendered and reference pixel colors, and $\mathcal{L}$ is the combination of $\mathcal{L}_1$ and SSIM losses in 3DGS. Our main objective is to utilize 3DGS in the extremely sparse setting and avoid overfitting to the input images.

\begin{figure*}[t]
% \vspace{-0.1in}
  \centering
  \includegraphics[width=1.0\linewidth]{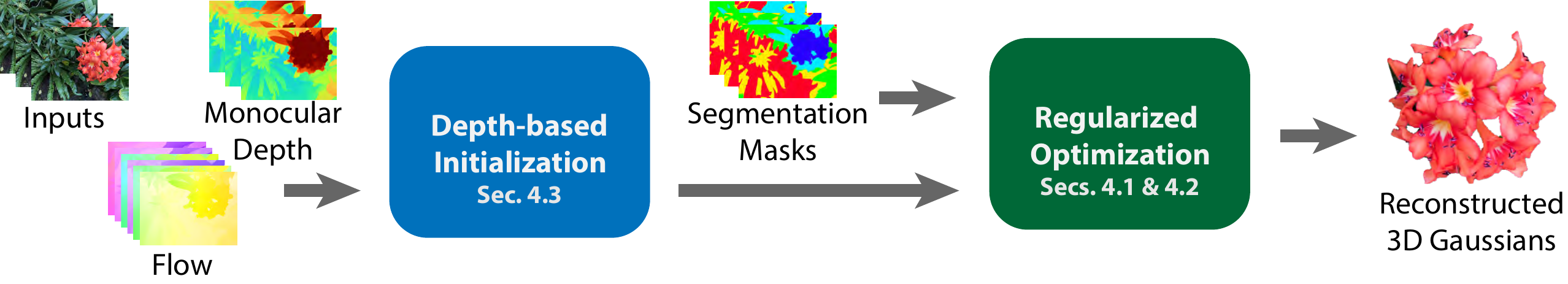}
  % \vspace{-10pt}
  \caption{\textbf{Overview of the optimization pipeline.} For every input image, we obtain monocular depth (Depth Anything~\cite{yang2024depth}) and dense flow correspondences between all image pairs (FlowFormer++~\cite{shi2023flowformer++}). These inputs are utilized to initialize a good set of 3D Gaussians for the subsequent optimization stage. The initialized 3D Gaussians, along with depth-based segmentation masks, are then used to perform a regularized 3D Gaussian optimization to obtain high-quality reconstruction.}
  \label{fig:overview}
  % \vspace{-10pt}
\end{figure*}

\section{Algorithm}
\label{sec:algorithm}

Given a sparse set of $N$ images (e.g., 3 or 4 images), our goal is to reconstruct a 3D Gaussian representation of the static scenes. Our key idea is to introduce coherency to the 3D Gaussians during optimization. In other words, when the position of a Gaussian is updated, the neighboring Gaussians should also be similarly affected during optimization. With this coherency, we can then use sparse regularization to further constrain the optimization and avoid overfitting to the input images. The overview of our method is shown in Fig.~\ref{fig:overview}. We discuss our approach in detail in the following sections.

\subsection{Coherent 3D Gaussian Optimization}

Introducing coherency to unstructured Gaussian particles in 3D is challenging. Our main idea is to transform the representation to a more structured form in 2D image domain. To do this, we propose to assign a single Gaussian to each pixel in every input image. We further restrict the movement of the Gaussians at each pixel to a ray connecting the center of the camera to that pixel. Under this representation, the position of each Gaussian can be controlled using a scalar depth value. Specifically, given an initial depth estimate at each pixel (see Sec.~\ref{sec:3D Gaussian Initialization}), we update the position of each Gaussian through residual depth as follows: 

% \vspace{-0.1in}
\begin{equation}
\label{eqn:pixel_gauss}
\mathbf{x} = g(D^{\text{init}}_n [\mathbf{p}] + \Delta D_n [\mathbf{p}], \mathbf{p}),
\end{equation}
% \vspace{-0.1in}

\noindent where the function $g(d, p)$ projects pixel $p$ into 3D according to depth $d$. Assuming that the local surface geometry is reasonably captured by the initial depth map at each view $D^{\text{init}}_n$, the residual depth should only vary smoothly to adjust the inaccuracies. Based on this observation, we propose single-view and multiview constraints as discussed below.

\begin{figure}[t]
  \centering
  \includegraphics[width=0.75\linewidth]{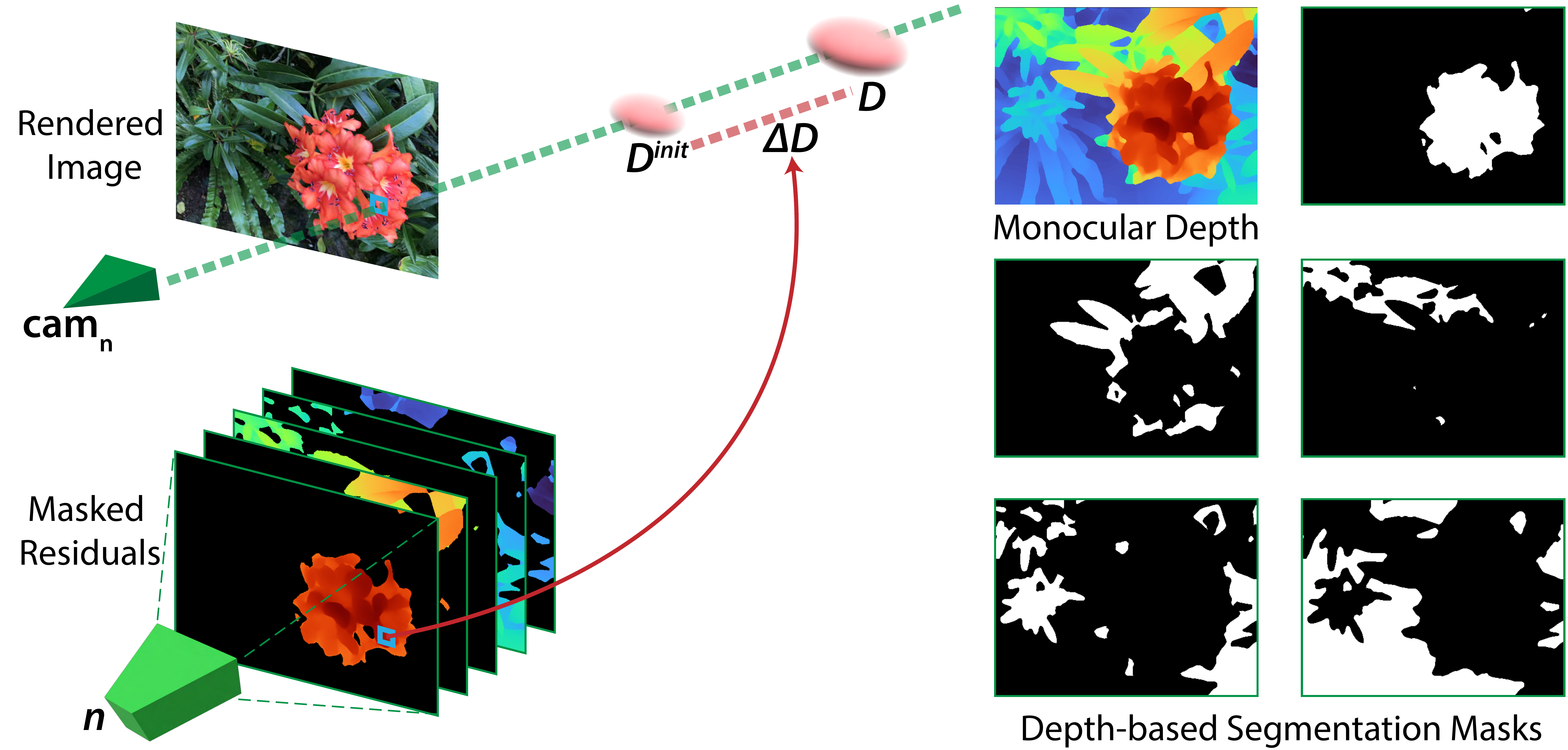}
  % \vspace{-0.1in}
  \caption{During regularized optimization, the implicit decoder
predicts the residual depth $\Delta D$ that moves the Gaussians from their initial position towards the true scene depth $D$. The input coordinate $n$ to the decoder corresponds to the input view with camera $cam_n$. To preserve sharp discontinuities, we apply binary segmentation masks to the decoder output obtained by thresholding the monocular depth.}
  \label{fig:implicit_decoder}
  % \vspace{-0.1in}
\end{figure}

\textbf{Single-view Constraint}\quad We enforce per-view smoothness by utilizing an implicit convolutional decoder. Specifically, as shown in Fig.~\ref{fig:implicit_decoder}, this decoder takes the normalized view index $n$ as the input and estimates the residual depth for the entire image, i.e., $\Delta D_n = f_{\phi}(n)$, where $\phi$ refers to the parameters of the decoder. Instead of updating the residual depth at individual pixels, we perform the optimization by modifying the decoder parameters $\phi$. This ensures that the residual depth is smooth and object surfaces are coherently deformed during optimization.

However, smooth deformation means that the decoder struggles to handle the sharp depth discontinuities between the objects. We address this by obtaining a $C$-channel binary segmentation mask, \camready{$S$}, through dividing each image into $C$ (5 in our implementation) separate regions, each with similar depths. Specifically, we use the approach by Yang et al.~\cite{yang2024depth} to estimate the monocular depth and follow the strategy by Wang et al.~\cite{wang20223d} to divide the image into $C$ regions based on the input depth (see Fig.~\ref{fig:implicit_decoder} - right). The decoder in this case, also produces a $C$-channel residual depth. The final residual is obtained as $\Delta D_n = \text{chsum}(S \odot f_{\phi}(n))$, where chsum refers to the channelwise summation operation.

We also apply a similar constraint to the global opacity $\alpha$ of the Gaussians to ensure coherency. In this case, we start from an initial opacity and the decoder estimates a $C$-channel residual opacity which is combined with the initial value to obtain the final opacity at each pixel.

\textbf{Multiview Constraint}\quad While the single-view constraint ensures smooth deformation of Gaussians corresponding to each image, it does not enforce the 3D surfaces, formed by the Gaussians from all the images, to be smooth. To encourage the reconstructed geometry to be smooth, we propose to utilize a total variation (TV) regularization $\mathcal{L}_\mathrm{TV}$ on the rendered depth. Specifically, we replace the color $\mathbf{c}$ with depth $d$ in Eq.~\ref{eq:rendered_color} to obtain the rendered depth $R_{\Sigma, \alpha, \mathbf{x}, d}$ in each view. We then apply a TV loss on the rendered disparity (inverse of depth) in two ways as follows:

% \vspace{-0.1in}
\begin{equation}
\label{eqn:tv_loss}
\mathcal{L}_\mathrm{TV} = \bigg\Vert \nabla 
\left(\frac{1}{1 + R_{\Sigma, \alpha, \mathbf{x}, d}}\right) \bigg\Vert_1, \quad \mathcal{L}_\mathrm{MTV} = \bigg\Vert \nabla \left(\mathbf{S}\odot
\left(\frac{1}{1 + R_{\Sigma, \alpha, \mathbf{x}, d}}\right)\right) \bigg\Vert_1.
\end{equation}
% \vspace{-0.1in}

Here, $\mathcal{L}_\mathrm{TV}$ enforces the depth to be overall smooth, while $\mathcal{L}_\mathrm{MTV}$ ensures that the depth in each segmented region is smooth. We propose to start by minimizing $\mathcal{L}_\mathrm{TV}$ to obtain a globally smooth and connected geometry (no sharp discontinuities) and gradually increase the contribution of $\mathcal{L}_\mathrm{MTV}$ to improve the structure details. We do this through the following loss:

\begin{equation}
\label{eqn:smooth_loss}
\mathcal{L}_\mathrm{multi} = (1 - \lambda_{s})\mathcal{L}_\mathrm{TV} + \lambda_{s}\mathcal{L}_\mathrm{MTV},
\end{equation}

\noindent where $\lambda_s$ is initialized to 0 and is gradually increased to reach 1 by the end of the optimization. With the coherency introduced to the 3D Gaussian optimization, we can now apply additional sparse regularization to improve the results as discussed below.

\begin{figure}[t]
% \vspace{-0.2cm}
  \centering
  \includegraphics[width=0.9\linewidth]{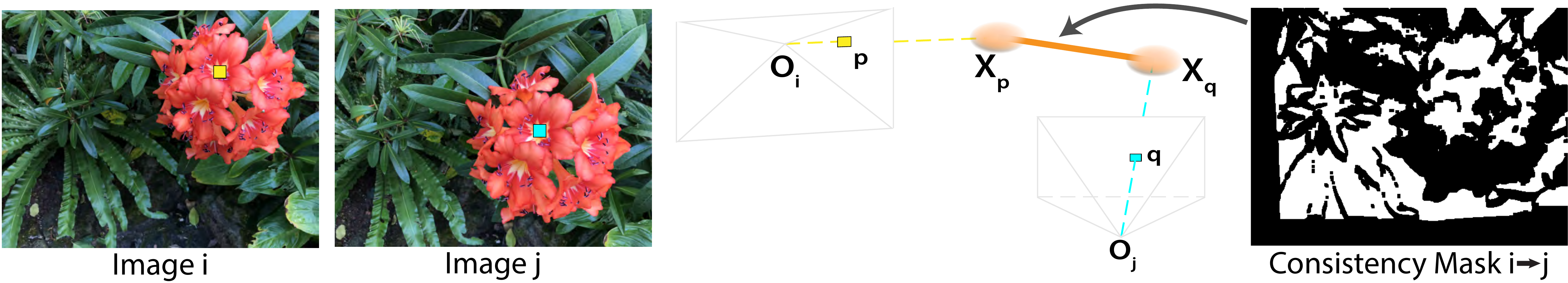}
  % \vspace{-10pt}
  \caption{Our flow based regularization forces the Gaussians of corresponding pixels in a pair of images (e.g.,
\textcolor{yellow}{yellow} and \textcolor{cyan}{cyan} squares) to have similar positions, by minimizing their \textcolor{orange}{distance}. The binary mask is utilized to mask out unreliable correspondences.}
  \label{fig:3d_warping}
  % \vspace{-0.2in}
\end{figure}

\subsection{Additional Regularization} 
\label{sec:Sparse Regularization}

Inspired by consistent depth estimation techniques~\cite{luo2020consistent, kopf2021robust, zeng2023test}, we propose a flow-based regularization term to further constrain the problem, as shown in Fig.~\ref{fig:3d_warping}. Specifically, the key idea is that the corresponding points in two input images come from the same 3D point. Therefore, we force the position of the Gaussians of the corresponding pixels in two images to be similar. This can be formally written as:

% \vspace{-0.1in}
\begin{equation}
\label{eqn:tv_loss1}
\mathcal{L}_\mathrm{flow} = \sum_{(i, j)}\sum_{\mathbf{p}} \bigg\Vert M_{i \rightarrow j} \odot \bigg( g(D_i [\mathbf{p}], \mathbf{p}) - g(D_j [\mathbf{q}], \mathbf{q}) \bigg) \bigg\Vert_1
\end{equation}
% \vspace{-0.1in}

\noindent where $D_i [\mathbf{p}] = D^{\text{init}}_i [\mathbf{p}] + \Delta D_i[\mathbf{p}]$, and $g(d, p)$ is a function that projects pixel $p$ into the 3D space according to its depth $d$. Moreover, $\mathbf{p}$ and $\mathbf{q}$ are the corresponding pixels in cameras $i$ and $j$, respectively, and are calculated using an existing optical flow method (Shi et al.~\cite{shi2023flowformer++} in our implementation). Furthermore, $M_{i \rightarrow j}$ is a binary mask, indicating the reliable correspondences, obtained using the forward-backward consistency check~\cite{luo2020consistent}.

In summary, our proposed coherent 3D Gaussian optimization with the additional regularization is performed by minimizing the following objective:

% \vspace{-0.1in}
\begin{equation}
    \label{eq:3dgs_opt_reg}
	\Sigma^*, \phi^*, \mathbf{c}^* = \arg \min_{\Sigma, \phi, \mathbf{c}}\sum_{\mathbf{p} \in \mathcal{P}}\,\mathcal{L}(R_{\Sigma, \alpha, \mathbf{x}, \mathbf{c}}(\mathbf{p}), R({\mathbf{p}})) + \beta_m \mathcal{L_\text{multi}} + \beta_f \mathcal{L_\text{flow}},
\end{equation}
% \vspace{-0.1in}

\noindent where $\beta_m$ and $\beta_f$ control the contribution of the multiview and flows terms and we set them to 5 and 0.1, respectively. Note that the global opacity $\alpha$ and position $\mathbf{x}$ of the Gaussians are optimized indirectly by updating the parameters of the implicit decoder $\phi$.

While 3DGS optimizes the objective by only sampling the center of each pixel, we found this strategy to be problematic for sparse input setting. In this case, the Gaussians will be deformed to match the color at the center of each pixel, leaving the remaining areas uncovered. As a result, the surfaces, when viewed from a novel view, will appear semi-transparent. To address this issue, we perform the optimization by evaluating multiple samples within each pixel \camready{and averaging them to obtain the pixel color}. The multisampling ensures that the Gaussians properly cover each pixel, resulting in significantly improved images, as shown in the supplementary video and Table~\ref{tab:ablation}.

\subsection{3D Gaussian Initialization} 
\label{sec:3D Gaussian Initialization}

\begin{figure}[t]
% \vspace{-0.2cm}
  \centering
  \includegraphics[width=0.8\linewidth]{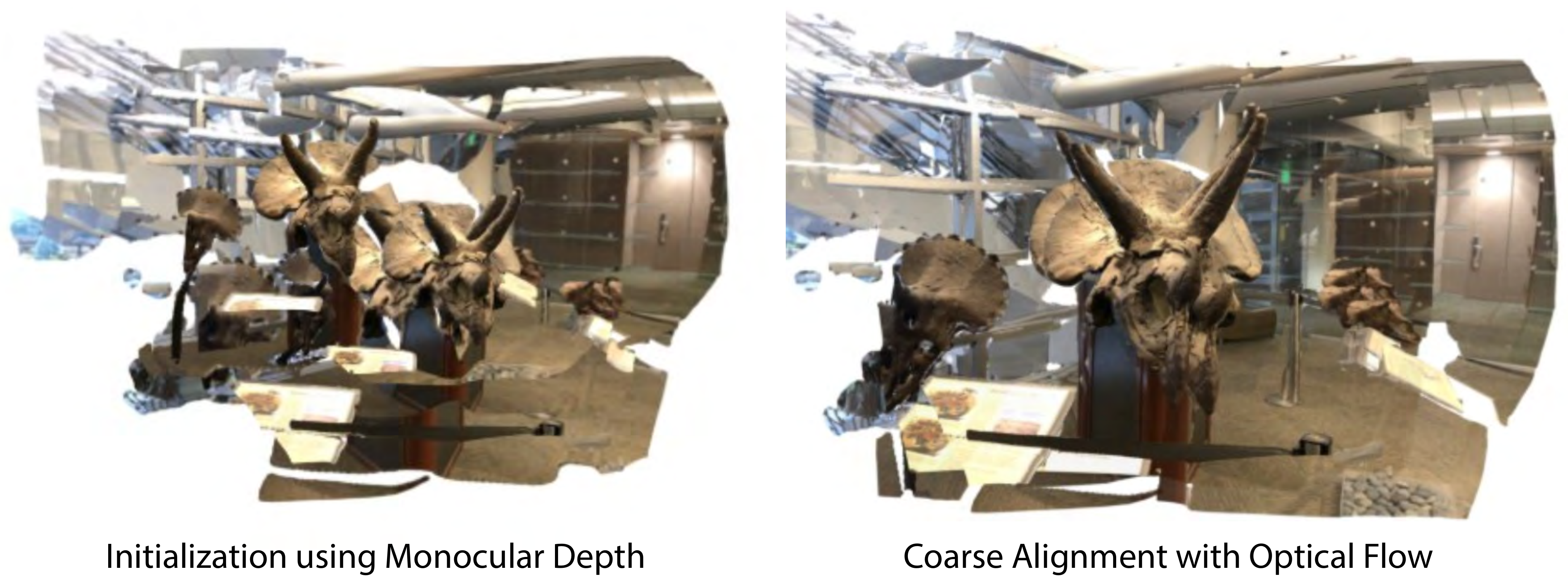}
  \vspace{-10pt}
  \caption{We initialize a set of 3D Gaussians from each view using monocular depth to support our regularized optimization. However, since the monocular depth is relative, the initialized representation is not multi-view consistent (left). Therefore, before Gaussian initialization, we coarsely align the representations from different images using flow correspondences (right). This ensures that the optimization begins from a sensible starting point, which proves to be essential for the training of 3D Gaussian representation under the challenging ill-conditioned setting.}
  \label{fig:pre_alignment}
  % \vspace{-0.2in}
\end{figure}

To facilitate our regularized optimization pipeline, we need a suitable initialization. In particular, our optimization requires an initial depth estimate $D^{\text{init}}$ that captures the local geometry of objects reasonably well. We do so using a monocular depth estimation method (Yang et al.~\cite{yang2024depth} in our implementation). While these approaches produce high-quality depth maps, the estimated depth is relative and often not consistent across different views. Therefore, by performing the initialization using these approaches, the Gaussians corresponding to the same surfaces across different views exhibit significant misalignments, as shown in Fig.~\ref{fig:pre_alignment} (left), hindering the optimization process.

To mitigate this issue, we use a flow-based loss, similar to the one described in Sec.~\ref{sec:Sparse Regularization}, to optimize the monocular depth. Directly optimizing the depth, however, could be problematic as the loss is only enforced in the areas where the flow is accurate, leaving the depth corresponding to remaining areas unaltered. Therefore, we propose to only optimize the scale and offset of the depth at each image by minimizing the following objective:

% \vspace{-0.1in}
\begin{equation}
\label{eqn:coarse}
\mathbf{s}^*, \mathbf{o}^* = \arg \min_{\mathbf{s}, \mathbf{o}} \sum_{(i,j)}\sum_{\mathbf{p}}\bigg\Vert M_{i \rightarrow j} \odot \bigg( g(s_i \cdot D^{\text{m}}_i[\mathbf{p}] + o_i, \mathbf{p}) - g(s_j \cdot D^{\text{m}}_j[\mathbf{q}] + o_j, \mathbf{q}) \bigg) \bigg\Vert_1,
\end{equation}
% \vspace{-0.1in}

\begin{wrapfigure}{r}{0.4\textwidth}%[t]
% \vspace{-0.2cm}
  \centering
  \includegraphics[width=0.4\textwidth]{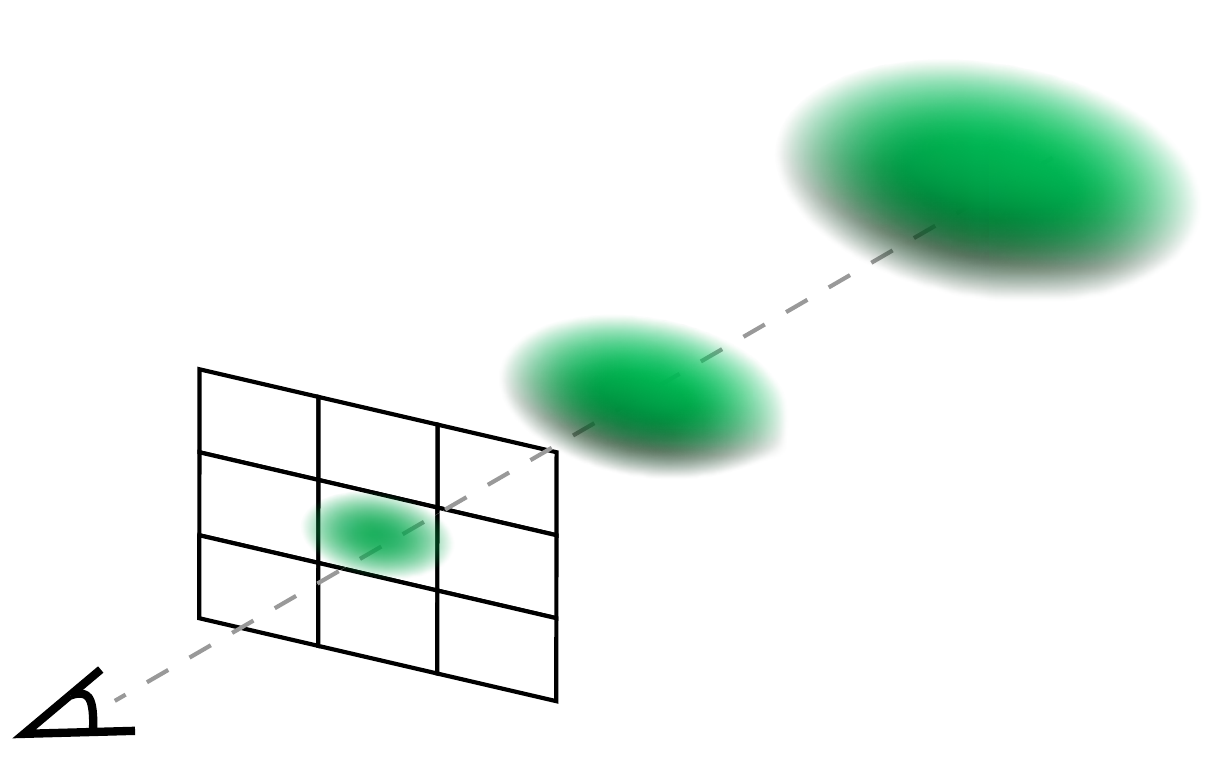}
  % \vspace{-10pt}
  \caption{Gaussian Scaling}
  \label{fig:depth_scaling}
  % \vspace{-0.1in}
\end{wrapfigure}
\noindent where $D^{\text{m}}_i$ is the monocular depth at camera $i$. Once the optimization is complete, we obtain the optimal scale and offset for each depth map. Our initial depth can then be obtained by applying the scale and offset to the monocular depth, i.e., $D^{\text{init}} = s\cdot D^{\text{m}} + o$. The result of this alignment can be observed in Fig.~\ref{fig:pre_alignment} (right).

\camready{An alternative approach is to utilize multi-view stereo (MVS)~\cite{yao2018mvsnet, zhe2023geomvsnet, xiong2023cl} which provides a 3D consistent alternative to monocular depth supervision, as shown by MVSNeRF~\cite{chen2021mvsnerf}. However, the state-of-the-art MVS approaches generate depth with significant artifacts for sparse images with large baseline, which hinders the optimization. High quality monocular depth better complements our decoder-based approach to obtain a coherent representation.}

Following 3DGS, we represent the covariance matrix in terms of rotation and scale matrices. We initialize the rotation matrix to identity. For the scale matrix, we use isometric scale and treat the Gaussians as spheres. We then compute the scale according to the initial depth such that each Gaussian covers its corresponding pixel properly, as shown in Fig.~\ref{fig:depth_scaling}. This is done as follows:

% \vspace{-0.1in}
\begin{equation}
\label{eqn:radius}
r = {f \cdot D^{\text{init}}}/{H},
\end{equation}
% \vspace{-0.1in}

\noindent where $r$ is the radius of the sphere and is used to create the isometric scale matrix, and $f$ represents the vertical focal length of the input image with height $H$. For the color, represented in terms of the SH coefficients, we set the DC value to the pixel color and other coefficients to zero. \camready{The initial opacity of gaussians is set based on the number of views, $2=0.6, 3=0.5, 4=0.35$ across all datasets. We decrease the initial opacity with the number of views to ensure gradient propagation and better alignment.}

\section{Experiments}

We implement our technique in PyTorch and use Adam~\cite{kingma15adam} for optimization. We perform the coarse alignment during initialization for 1,000 iterations and the regularized optimization for 13,000 iterations. During the first 8,000 iterations of the regularized optimization, we keep the rotation matrix as identity and assign the scale according to Eq.~\ref{eqn:radius}. For the remaining 5,000 iterations, we freely optimize both the rotation and scale using the objective function.

In the following sections, we provide quantitative and qualitative comparisons against state-of-the-art approaches and evaluate the effect of various components of our method.

\subsection{Comparisons}
\label{sec:results}
We compare our approach against the state-of-the-art NeRF-based approaches for sparse view synthesis. We specifically compare with approaches by Yang et al.~\cite{yang2023freenerf} (FreeNeRF), Seo et al.~\cite{seo2023flipnerf} (FlipNeRF), Wang et al.~\cite{wang2022sparsenerf} (SparseNeRF) and Niemeyer et al.~\cite{Niemeyer2021Regnerf} (RegNeRF). Moreover, the vanilla 3DGS serves as an additional baseline. For all the approaches, we utilize the implementations provided by the authors. We showcase both the quantitative and qualitative performances on two datasets, LLFF~\cite{mildenhall2020nerf} and NVS-RGBD~\cite{wang2022sparsenerf}. We follow previous methods to divide the images in both datasets into training and test views.

\begin{table}[t]
\caption{Numerical comparisons on the LLFF~\cite{mildenhall2020nerf} dataset with 2 to 4 views.}
\centering
\begin{tabular}{l|ccc|ccc|ccc} \toprule
\multirow{2}{*}{\textbf{Methods}} & \multicolumn{3}{c|}{PSNR} & \multicolumn{3}{c|}{SSIM} & \multicolumn{3}{c}{LPIPS} \\ 
 \cline{2-10} 
 & 2 & 3 & 4 
 & 2 & 3 & 4                
 & 2 & 3 & 4                \\ \hline \hline

3DGS       & 12.83 & 14.99 & 17.31 & 0.311 & 0.483 & 0.584 & 0.470 & 0.362 & 0.297\\
RegNeRF    & 16.55 & 19.41 & 21.49 & 0.468 & 0.627 & 0.713 & 0.417 & 0.306 & \cellcolor{top2}0.257\\
FlipNeRF   & 16.57 & \cellcolor{top3}19.74 & 21.55 & \cellcolor{top3}0.485 & \cellcolor{top2}0.668 & \cellcolor{top2}0.721 & 0.407 & \cellcolor{top3}0.282 & 0.260\\
FreeNeRF   & \cellcolor{top3}17.07 & \cellcolor{top2}19.97 & \cellcolor{top2}21.80 & \cellcolor{top2}0.513 & 0.652 & 0.713 & \cellcolor{top2}0.376 & \cellcolor{top2}0.280 & \cellcolor{top3}0.259\\
SparseNeRF & \cellcolor{top2}17.74 & \cellcolor{top1}20.33 & \cellcolor{top1}21.90 & \cellcolor{top2}0.513 & \cellcolor{top3}0.657 & \cellcolor{top3}0.720 & \cellcolor{top3}0.386 & 0.302 & 0.260\\
Ours       & \cellcolor{top1}18.32 & \cellcolor{top1}20.33 & \cellcolor{top3}21.58 & \cellcolor{top1}0.644 & \cellcolor{top1}0.725 & \cellcolor{top1}0.762 & \cellcolor{top1}0.220 & \cellcolor{top1}0.180 & \cellcolor{top1}0.167\\
\bottomrule
\end{tabular}
% \vspace{-0.3em}
\label{tab:llff}
% \vspace{1em}
\end{table}

\textbf{LLFF Dataset} \quad We begin by showing the quantitative comparisons in terms of PSNR, SSIM, and LPIPS in Table~\ref{tab:llff}. Since our approach constrains the movement of the Gaussians to a ray connecting the camera center and the corresponding pixel, the regions that are occluded in all the views will not be reconstructed. To have a fair comparison against the other methods, we identify the occluded areas and mask them out in numerical evaluation. We use the same mask for all the approaches. To identify these regions, we render the opacity and consider any region with opacity below $1e-3$ as occluded. As seen in Table~\ref{tab:llff}, in most cases, our method produces better results across all the metrics. In particular, the perceptual quality of our results, measure by LPIPS, is significantly better than the other techniques in all the input settings.

\begin{figure}[!t]
% \vspace{-0.2in}
  \centering
  \includegraphics[width=0.9\linewidth]{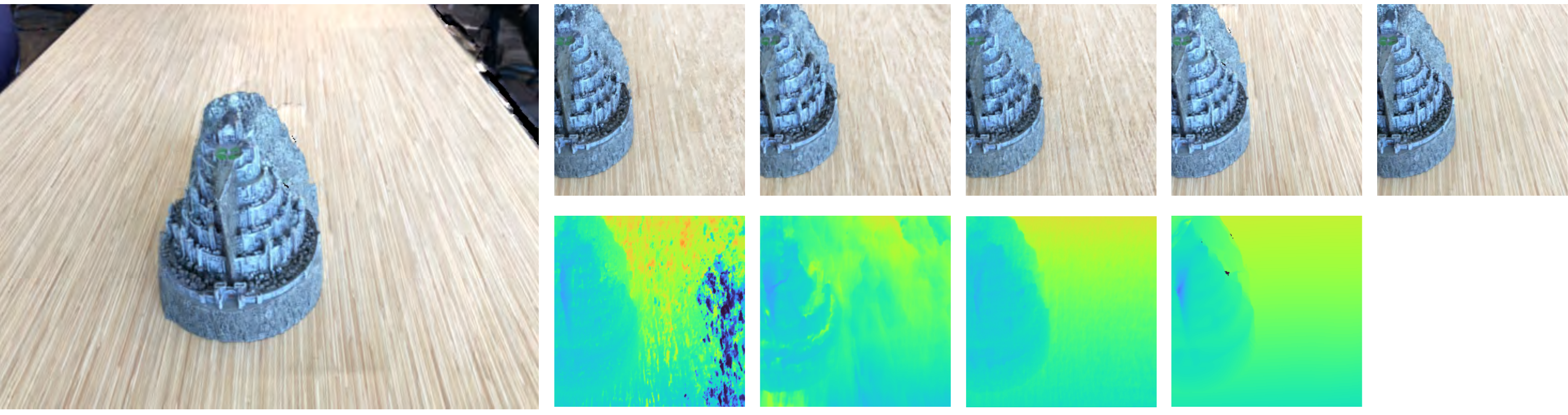}
  \includegraphics[width=0.9\linewidth]{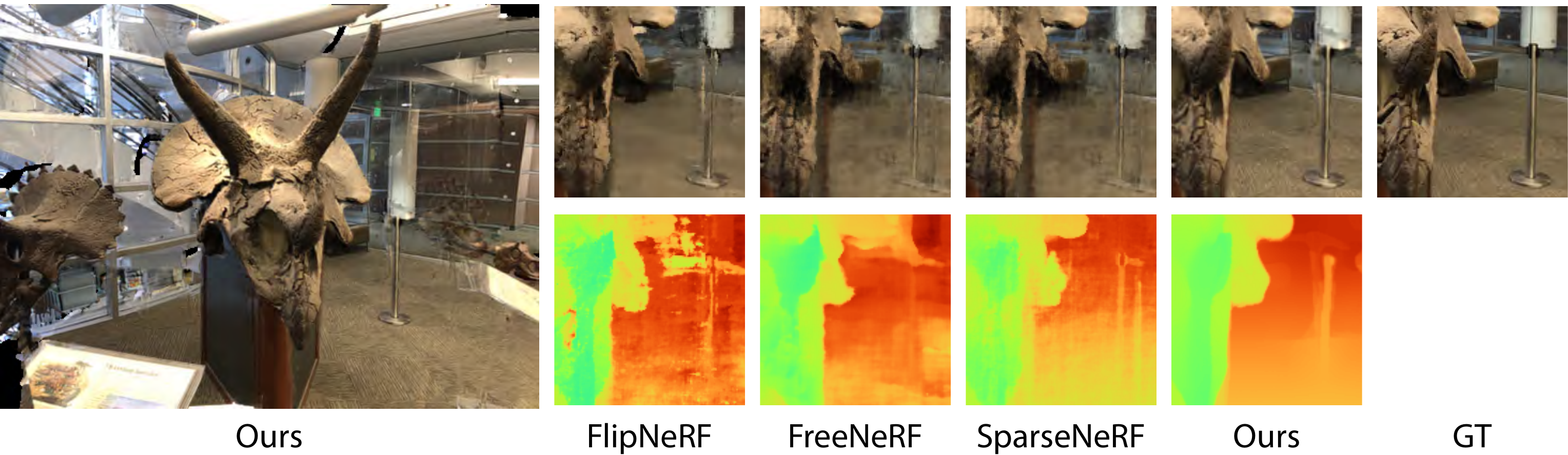}
  % \vspace{-10pt}
  \caption{\textbf{LLFF 3 input}. We show comparisons against other sparse-view NeRF-based approaches, SparseNeRF~\cite{wang2022sparsenerf}, FlipNeRF~\cite{seo2023flipnerf} and FreeNeRF~\cite{yang2023freenerf}. Our approach produces high-quality novel views while reconstructing significantly better geometry.}
  \label{fig:llff3}
  % \vspace{-10pt}
\end{figure}

Furthermore, we visually compare our approach against a subset of other methods on two scenes from the LLFF dataset with 3 input images. As shown in Fig.~\ref{fig:llff3}, our method significantly enhances the reconstruction of both texture and geometry compared to NeRF-based techniques. For the \textsc{Fortress} scene, both FlipNeRF and FreeNeRF generate noticeable artifacts and fail to accurately capture the underlying shape. SparseNeRF appears to generate an adequate depth, but upon closer examination, noisy artifacts in both texture and depth become apparent. Our method, on the other hand, is capable of creating high-quality texture with a clean and smooth depth. The second scene, \textsc{Horns}, is complex with multiple objects with intricate details at various depths. All the NeRF-based approaches introduce significant artifacts in the rendered results. In contrast, our method is able to generate a coherent and high-quality texture and geometry. Note that since our approach is based on 3DGS, our inference speed is significantly higher than the alternatives. On an Nvidia Tesla V100 GPU, our approach has an average inference speed of 278 fps for LLFF data with 3 inputs, while the other approaches perform rendering at 0.08 fps.

As discussed previously, unlike other methods that fill in the occluded regions with blurry and repetitive textures, our method does not reconstruct these areas. This gives us a unique advantage as we can identify and inpaint these areas to hallucinate high-quality details. As a proof of concept, we use a simple strategy to inpaint these areas with a diffusion model and project them to 3D using monocular depth \camready{(see supplementary Sec.2 for more details)}. As shown in Fig.~\ref{fig:3dgs_comparison} (right) the hallucinated details are of high-quality and consistent (see supplementary video).

\begin{table}[t]
\caption{Numerical comparisons on the NVS-RGBD~\cite{wang2022sparsenerf} dataset with 2 and 3 views.}
\centering
\begin{tabular}{l|cc|cc|cc|cc|cc|cc} \toprule
\multirow{3}{*}{\textbf{Methods}} & \multicolumn{6}{c|}{ZED 2} & \multicolumn{6}{c}{iPhone} \\ \cline{2-13}
 & \multicolumn{2}{c|}{PSNR} & \multicolumn{2}{c|}{SSIM} & \multicolumn{2}{c|}{LPIPS} & \multicolumn{2}{c|}{PSNR} & \multicolumn{2}{c|}{SSIM} & \multicolumn{2}{c}{LPIPS} \\ 
 \cline{2-13} 
 & 2 & 3 & 2 & 3 
 & 2 & 3 & 2 & 3                
 & 2 & 3 & 2 & 3                \\ \hline \hline

3DGS       & 13.51 & 14.43 & 0.313 & 0.315 & 0.545 & 0.545 & 12.66 & 14.22 & 0.330 & 0.382 & 0.563 & 0.531\\
RegNeRF    & 18.90 & \cellcolor{top3}25.56 & 0.580 & 0.803 & 0.352 & 0.169 & 16.04 & \cellcolor{top3}22.19 & 0.520 & 0.708 & 0.420 & \cellcolor{top2}0.271\\
FlipNeRF   & 17.28 & \cellcolor{top3}25.56 & 0.565 & \cellcolor{top3}0.817 & 0.402 & 0.221 & 17.68 & 21.92 & \cellcolor{top3}0.581 & \cellcolor{top3}0.714 & \cellcolor{top3}0.386 & 0.293\\
FreeNeRF   & \cellcolor{top3}20.23 & \cellcolor{top2}25.60 & \cellcolor{top3}0.689 & \cellcolor{top3}0.817 & \cellcolor{top2}0.265 & \cellcolor{top3}0.166 & \cellcolor{top1}18.33 & \cellcolor{top1}22.73 & \cellcolor{top2}0.598 & \cellcolor{top2}0.723 & \cellcolor{top2}0.372 & 0.281\\
SparseNeRF & \cellcolor{top2}22.07 & \cellcolor{top1}26.56 & \cellcolor{top2}0.694 & \cellcolor{top2}0.835 & \cellcolor{top3}0.285 & \cellcolor{top2}0.154 & \cellcolor{top3}17.87 & \cellcolor{top2}22.50 & 0.557 & \cellcolor{top1}0.725 & 0.409 & \cellcolor{top3}0.275\\
Ours       & \cellcolor{top1}23.05 & 24.93 & \cellcolor{top1}0.774 & \cellcolor{top1}0.840 & \cellcolor{top1}0.164 & \cellcolor{top1}0.135 & \cellcolor{top2}18.07 & 21.99 & \cellcolor{top1}0.615 & \cellcolor{top1}0.725 & \cellcolor{top1}0.292 & \cellcolor{top1}0.228\\
\bottomrule
\end{tabular}
% \vspace{-0.3em}
\label{tab:rgbd}
% \vspace{1em}
\end{table}

\begin{figure}[!t]
% \vspace{-0.2in}
  \centering
  \includegraphics[width=0.9\linewidth]{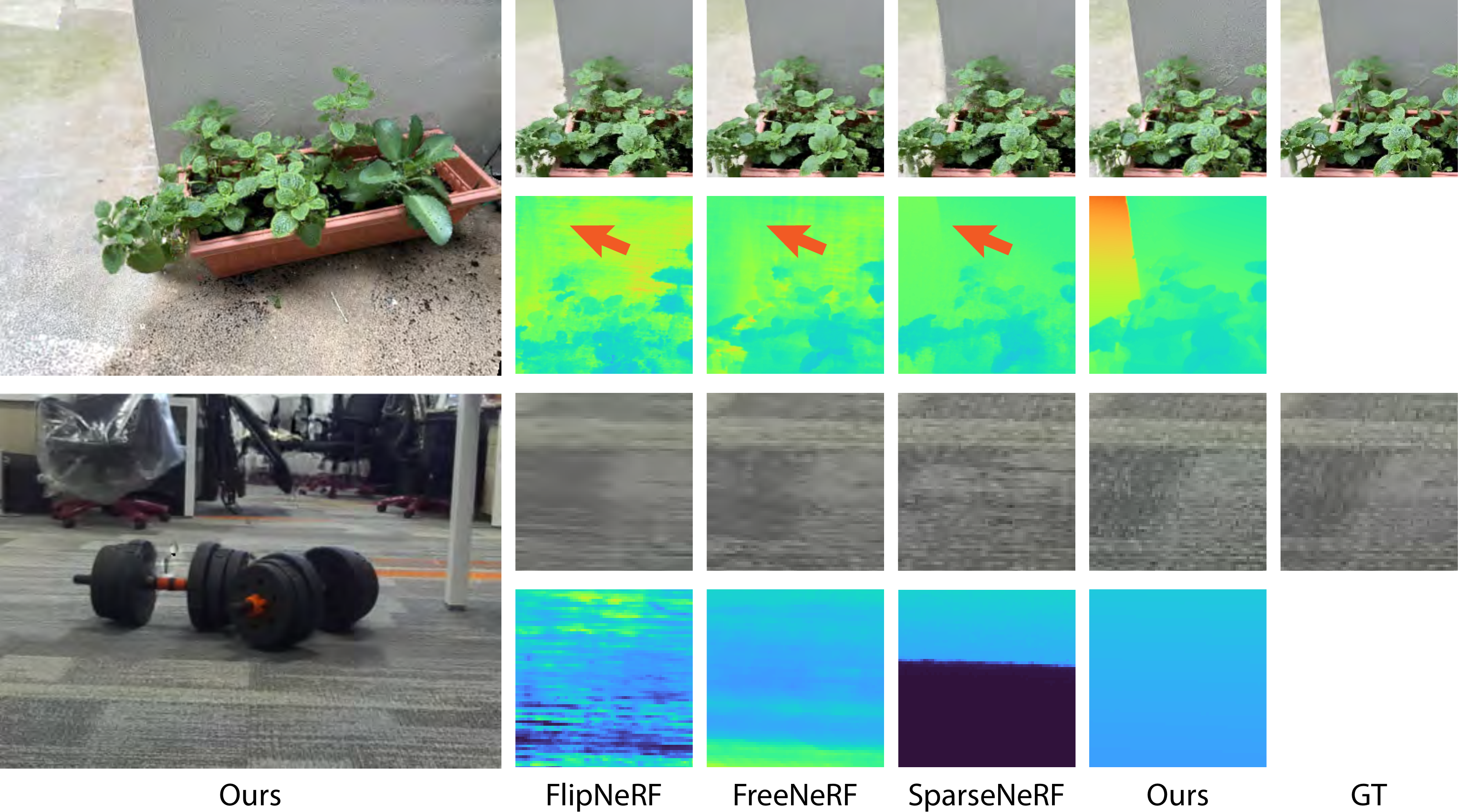}
  % \vspace{-10pt}
  \caption{\textbf{NVS-RGBD 3 input (iPhone on the top and ZED 2 on the bottom)}. We show comparisons against other sparse-view NeRF-based approaches. Our approach produces better texture details on the gray wall (top) and carpet (bottom) while reconstructing significantly better geometry.}
  \label{fig:rgbd3}
  % \vspace{-10pt}
\end{figure}

\textbf{NVS-RGBD Dataset}\quad Next, we numerically compare our method against the other approaches on NVS-RGBD dataset (ZED 2 and iPhone) with 2 and 3 input images. Again our approach produces overall better results than the other methods, particularly in terms of SSIM and LPIPS. The advantage is even larger with 2 input images.

Moreover, we provide visual results on two scenes with 3 inputs from the NVS-RGBD dataset in Fig.~\ref{fig:rgbd3}. For the top scene, NeRF-based approaches are not able to reconstruct the texture details on the gray wall. Additionally, none of the other methods are able to reconstruct the geometry of the floor properly, as indicated by the arrows. Our approach, on the other hand, clearly distinguishes the wall and floor and produces results with detailed texture. Similarly, other techniques fail to properly reconstruct the detailed texture of the carpet for the bottom scene. However, our method produces results that are close to ground truth with a smooth geometry.

\begin{figure}[!t]
%\vspace{-0.2cm}
  \centering
  \includegraphics[width=0.9\linewidth]{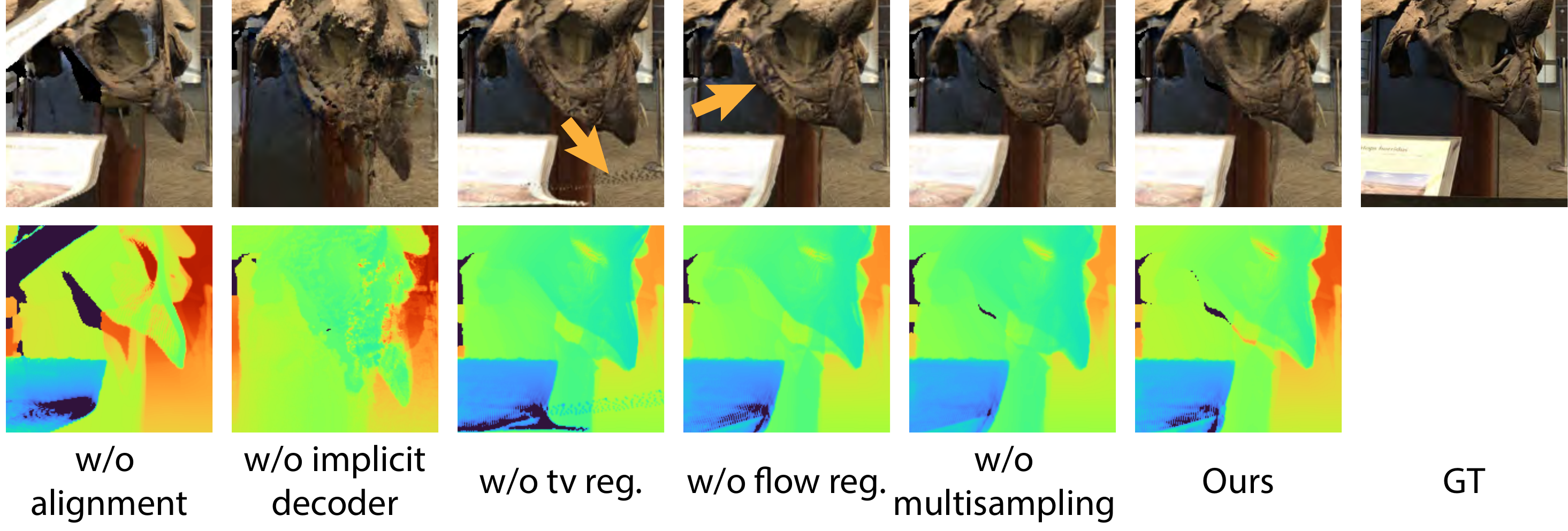}
  % \vspace{-10pt}
  \caption{We visually compare the effect of different components on the reconstruction quality. As seen, all the components are necessary to achieve high-quality results. The effect of multisampling might be difficult to see here and we encourage the readers to see our supplementary video.}
  \label{fig:ablation}
  % \vspace{-0.1in}
\end{figure}

\subsection{Ablations}
Here, we evaluate the effect of various components of our method on the reconstruction quality, both visually (Fig.~\ref{fig:ablation}) and numerically (Table.~\ref{tab:ablation}). The numerical results are obtained on the LLFF dataset with 3 input views. Without the flow-based coarse alignment (Eq.~\ref{eqn:coarse}) in the initialization stage, the optimization is unable to align the representation from different images for the complex scenes. During the regularized optimization process, the implicit decoder plays a critical role and ensures the reconstructed geometry is smooth. Furthermore, TV regularization suppresses the isolated point clouds that are not visible from the training views, but appear in a novel view, as indicated by the arrow. Moreover, without the flow-based regularization our approach has difficulty properly reconstructing the texture details in the unconstrained areas, as indicated by the arrow. Finally, multisampling improves the pixel coverage and avoids producing surfaces that appear as semi-transparent from the novel views. While this is difficult to see in Fig.~\ref{fig:ablation}, we encourage the readers to see the supplementary video where this effect is clearly visible.

\begin{table}[t]
\caption{Numerical comparisons to highlight the contribution of each component during the optimization on the LLFF dataset with 3 views.}
% \vspace{-0.15in}
\label{tab:ablation}
\begin{minipage}{\columnwidth}
\begin{center}
\begin{tabular}{lccc}
  \toprule
  Method & PSNR & SSIM & LPIPS\\
  \toprule
  \midrule
  w/o alignment           & 19.06 & 0.679 & 0.217\\ \hline
  w/o implicit decoder    & 16.68 & 0.477 & 0.331\\
  w/o tv reg.             & 20.20 & 0.724 & 0.186\\
  w/o flow reg.           & 20.32 & 0.723 & 0.185\\
  w/o multisampling       & 19.99 & 0.718 & 0.194\\ \hline
  Ours                    & \textbf{20.33} & \textbf{0.725} & \textbf{0.180}\\
  
  \bottomrule
\end{tabular}
\end{center}
\bigskip\centering
\end{minipage}
% \vspace{-0.25in}
\end{table}%

\section{Conclusion}
\label{sec:conclusion}
In this paper, we present a novel approach to regularize the 3DGS optimization for the sparse input setting. Specifically, we propose to assign a single Gaussian to every pixel of the input images to be able to constrain the Gaussians in 2D image space. We introduce coherency to the 3D Gaussian optimization pipeline using single and multi-view constraints through an implicit decoder and a total variation loss, respectively. We use monocular depth and flow correspondences to initialize a set of per-pixel 3D Gaussians to support the regularized optimization. This enables us to learn high quality texture and smooth geometry in the extreme sparse input setting. We demonstrate the superiority of our approach on various scenes from multiple datasets.

\begin{figure}[!t]
%\vspace{-0.2cm}
  \centering
  \includegraphics[width=1.0\linewidth]{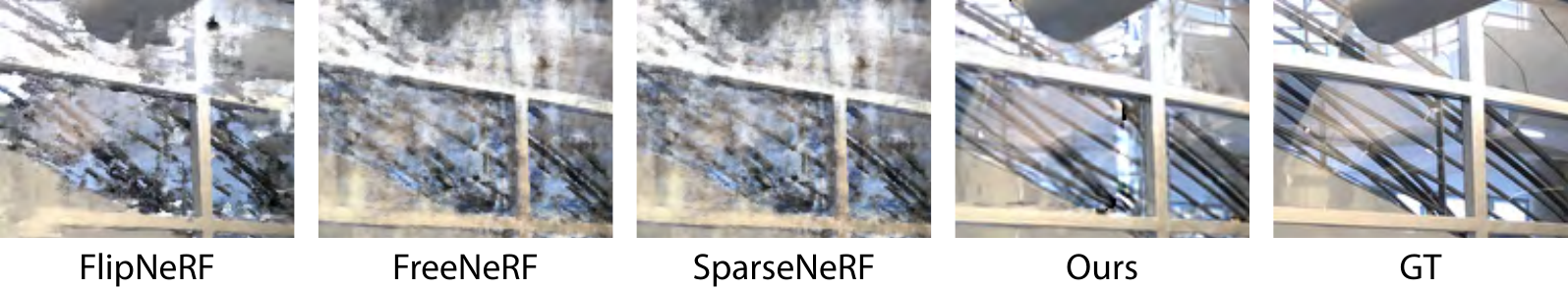}
  % \vspace{-10pt}
  \caption{Our approach assigns a single Gaussian to each pixel and as such reconstructing both the hand rails and the glass pane is difficult. Nonetheless, we are still better than existing NeRF-based approaches.}
  \label{fig:limitations}
  % \vspace{-0.1in}
\end{figure}

% \paragraph{Limitations and future work.}
\textbf{Limitations} \quad  Since we assign a single Gaussian to each pixel, our approach has difficulty handling scenes with transparent objects. An example of such a case is shown in Fig.~\ref{fig:limitations} where our technique is not able to properly reconstruct both the reflections on the glass and the hand rails behind it. Nevertheless, our results are still significantly better than the competing methods. Additionally, our approach relies on the monocular depth and may not be able to produce reasonable results if the depth is highly inaccurate.

\section*{Acknowledgements}
\camready{The project was funded in part by a generous gift from Meta.}

\bibliographystyle{splncs04}
\bibliography{egbib}

\clearpage

\begin{tightcenter}
\Large{\textbf{Supplementary Material}}
\end{tightcenter}

\section{Implicit Decoder Architecture}
We use the convolutional decoder network as proposed by Bemana et al.~\cite{bemana2020xfields, paliwal2023implicit} \camready{to output a full resolution 2D map corresponding to the camera index}. The \camready{normalized scalar} input index is concatenated to the coordconv~\cite{liu2018intriguing} layer in the first layer of the decoder. This is followed by a series of convolutional and upsampling layers (bi-linear interpolation) until the desired output resolution is reached. The number of parameters is controlled using a scalar capacity factor that is multiplied to a preset of each layer, e.g., [2c, 4c, 8c] where c is the capacity factor. We use capacity factor [10, 15, 18] for the depth decoder corresponding to 2, 3 and 4 inputs, respectively. Similarly, we use [6, 10, 12] for the opacity decoder.

\section{3D Inpainting}
\label{supp:3dinpainting}
\begin{figure}[!t]
% \vspace{-0.2in}
  \centering
  \includegraphics[width=0.9\linewidth]{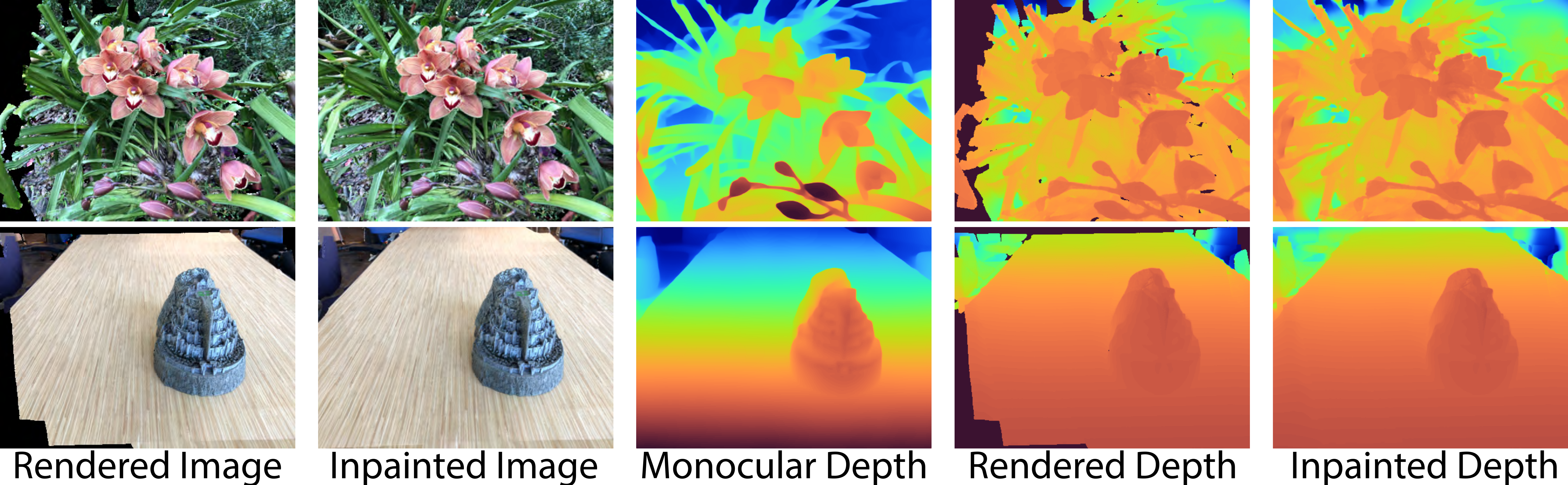}
  \vspace{-10pt}
  \caption{3D consistent scene inpainting results.}
  \label{fig:inpainting}
  % \vspace{-10pt}
\end{figure}

\camready{Since we constrain the movement of Gaussians, our approach, unlike existing methods, does not hallucinate details in the occluded areas. This is a unique advantage as it allows us to inpaint these regions using state-of-the-art methods. To do so, we begin by rendering a novel view with holes and alpha mask. We use stable diffusion inpainting~\cite{rombach2022high} to generate the missing texture in the masked regions. Then, we estimate the monocular depth on the inpainted image. Next, we use the gradient of this depth to fill in rendered depth in the masked regions using poisson blending~\cite{perez2003poisson}. We then project Gaussians to the scene using the inpainted depth and image. We repeat this process sequentially for a few novel views. Once this process is done, the inpainted scene is 3D consistent and can be rendered from any novel view without holes. We show additional inpainting results in Fig.~\ref{fig:inpainting}.}

\section{Comparison against DNGaussian}
\begin{figure}[!t]
% \vspace{-0.2in}
  \centering
  \includegraphics[width=0.9\linewidth]{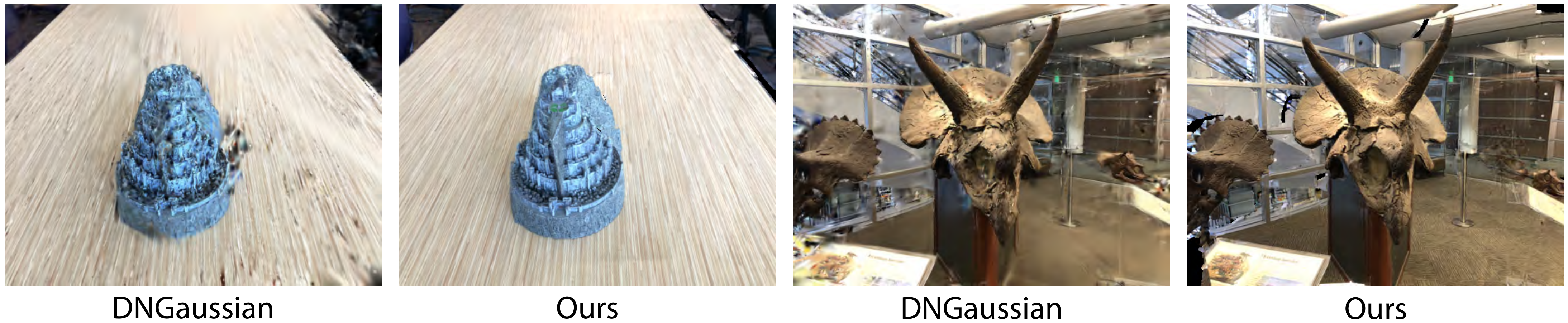}
  \vspace{-10pt}
  \caption{\textbf{LLFF 3 input}. We show comparisons against DNGaussian~\cite{li2024dngaussian}.}
  \label{fig:dngaussian}
  % \vspace{-10pt}
\end{figure}

\camready{We compare our approach against a recent state-of-the-art sparse view synthesis approach by Li et al.~\cite{li2024dngaussian} which also utilizes the 3D Gaussian representation. We use the author provided results for both qualitative (Fig.~\ref{fig:dngaussian}) and quantitative (Table \ref{tab:dngaussian}) comparisons on the 3-input LLFF~\cite{mildenhall2020nerf} dataset. We outperform DNGaussian both visually and numerically since DNGaussian generates blurry texture with distracting artifacts.}

\section{Additional Ablation Results}
\camready{To improve the effectiveness of decoder, we estimate a multi-channel depth offset by utilizing a multi-channel depth-based segmentation mask. Table.~\ref{tab:ablationmask} shows the effectiveness of this multi-channel approach.}

\begin{table}[t]
\caption{Numerical comparisons with DNGaussian~\cite{li2024dngaussian} on the LLFF dataset with 3 views.}
% \vspace{-0.15in}
\label{tab:dngaussian}
\begin{minipage}{\columnwidth}
\begin{center}
\setlength{\tabcolsep}{1em}
\begin{tabular}{lccc}
  \toprule
  Method & PSNR & SSIM & LPIPS\\
  \toprule
  \midrule
  DNGaussian & 19.55 & 0.647 & 0.264\\
  Ours & \textbf{20.33} & \textbf{0.725} & \textbf{0.180}\\
  
  \bottomrule
\end{tabular}
\end{center}
\bigskip\centering
\end{minipage}
% \vspace{-0.25in}
\end{table}%

\begin{table}[t]
\caption{Numerical comparisons to highlight the effect of \#channels in the segmentation mask.}
% \vspace{-0.15in}
\label{tab:ablationmask}
\begin{minipage}{\columnwidth}
\begin{center}
\setlength{\tabcolsep}{1em}
\begin{tabular}{lccc}
  \toprule
  Method & PSNR & SSIM & LPIPS\\
  \toprule
  \midrule
  Ours w/o mask                                 & 19.70 & 0.673 & 0.214\\
  Ours w/ 3ch mask                                 & 19.86 & 0.704 & 0.198\\
  Ours w/ 5ch mask & \textbf{20.33} & \textbf{0.725} & \textbf{0.180}\\
  Ours w/ 7ch mask                                 & 20.13 & 0.713 & 0.187\\
  
  \bottomrule
\end{tabular}
\end{center}
\bigskip\centering
\end{minipage}
% \vspace{-0.25in}
\end{table}%

\section{Additional Qualitative Results}
We provide additional qualitative results for 2 and 4 input configurations on the LLFF~\cite{mildenhall2020nerf} and NVS-RGBD~\cite{wang2022sparsenerf} datasets.

\begin{figure}[!t]
% \vspace{-0.2in}
  \centering
  \includegraphics[width=1.0\linewidth]{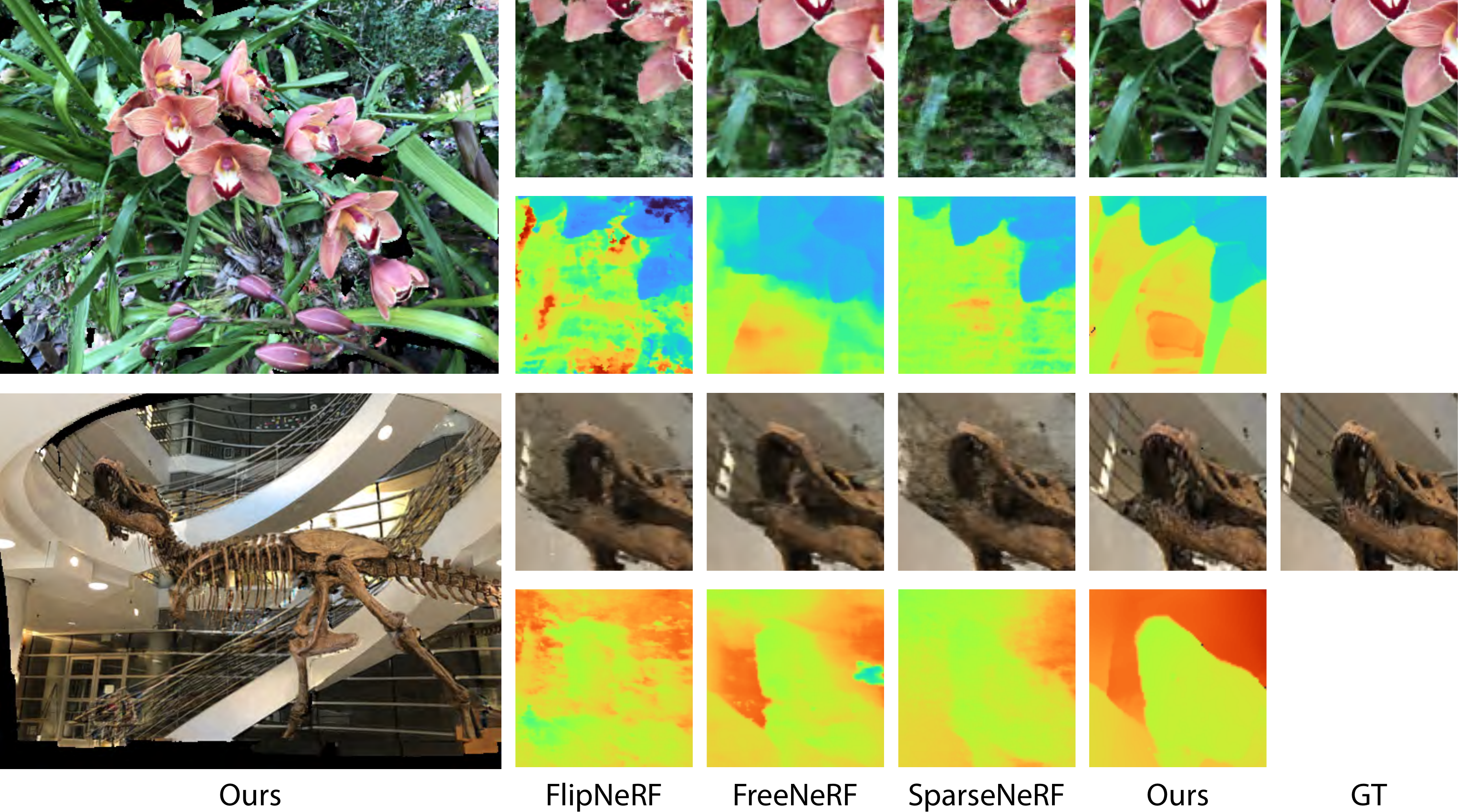}
  \vspace{-10pt}
  \caption{\textbf{LLFF 2 input}. We show comparisons against other sparse-view NeRF-based approaches, SparseNeRF~\cite{wang2022sparsenerf}, FlipNeRF~\cite{seo2023flipnerf} and FreeNeRF~\cite{yang2023freenerf}.}
  \label{fig:llff2}
  % \vspace{-10pt}
\end{figure}

\begin{figure}[!t]
% \vspace{-0.2in}
  \centering
  \includegraphics[width=1.0\linewidth]{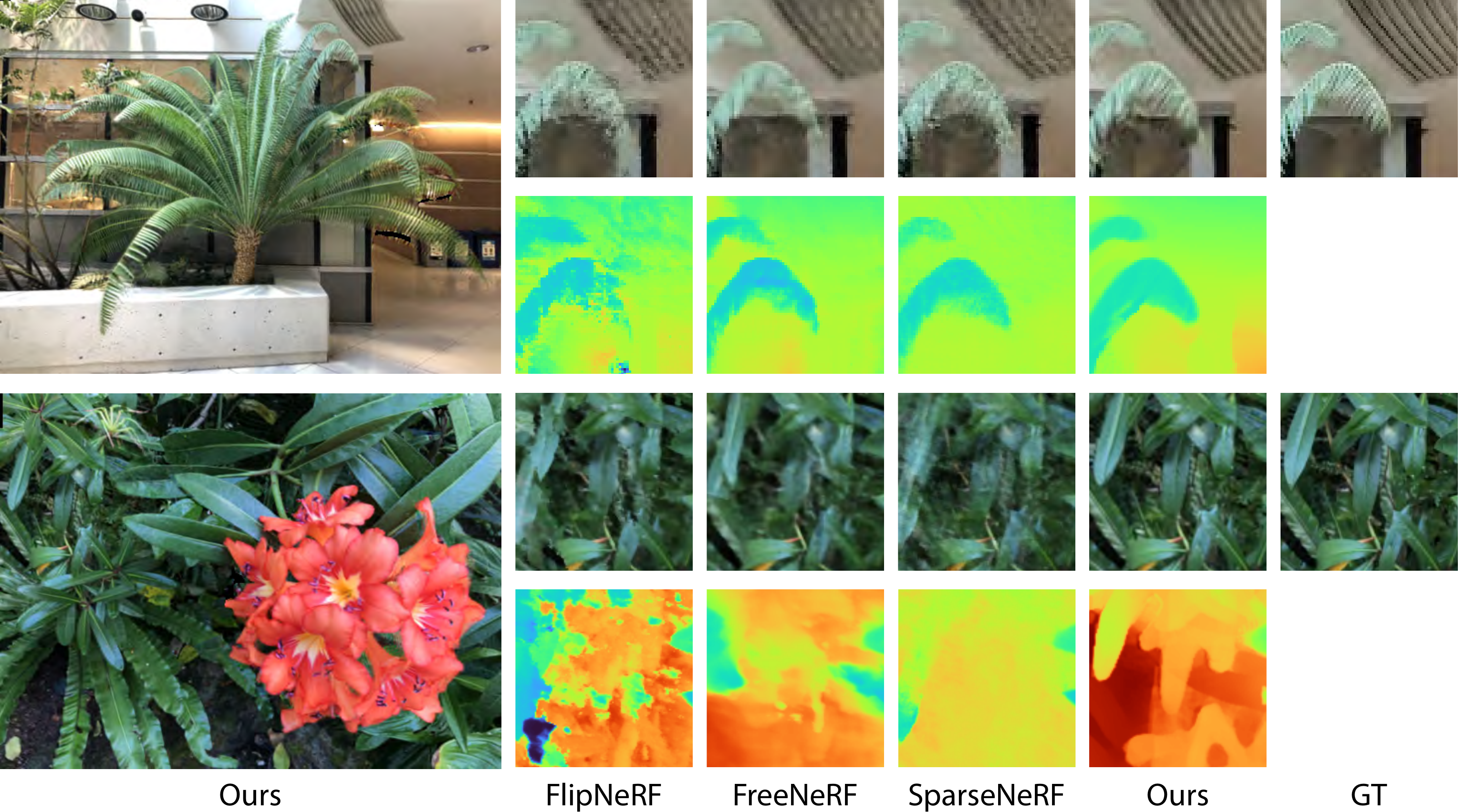}
  \vspace{-10pt}
  \caption{\textbf{LLFF 4 input}. We show comparisons against other sparse-view NeRF-based approaches, SparseNeRF~\cite{wang2022sparsenerf}, FlipNeRF~\cite{seo2023flipnerf} and FreeNeRF~\cite{yang2023freenerf}.}
  \label{fig:llff4}
  % \vspace{-10pt}
\end{figure}

\textbf{LLFF Dataset} \quad As shown in Fig.~\ref{fig:llff2}, our method produces significantly better texture and geometry compared to other approaches on the LLFF dataset with 2 inputs. For the Orchids scene, all the NeRF-based approaches fail to handle the complex jumble of leaves behind the flower, producing glaring artifacts. Our approach is able to capture the intricate geometry and details. The other scene, Trex, contains a lot of thin details as highlighted by the insets. NeRF-based approaches fail to capture the thin details and produce blurry texture on the trex bones. Our approach is able to reconstruct details while providing a smooth geometry.
We show the 4 input results in Fig.~\ref{fig:llff4}. SparseNeRF~\cite{wang2022sparsenerf} and FlipNeRF~\cite{seo2023flipnerf} produce noisy texture for the Fern scene, while FreeNeRF~\cite{yang2023freenerf} produces over-blurred results. Our approach produces texture much closer to ground truth in comparison without noise or blurriness. NeRF-based approaches struggle to handle regions that contain relatively sparse supervision (e.g., regions visible in 1 or 2 views out of 4). This is highlighted in the Flower scene. As shown, NeRF-based approaches produce ghosted artifacts while our method is able to generate a coherent geometry and thus high quality details.

\begin{figure}[!t]
% \vspace{-0.2in}
  \centering
  \includegraphics[width=1.0\linewidth]{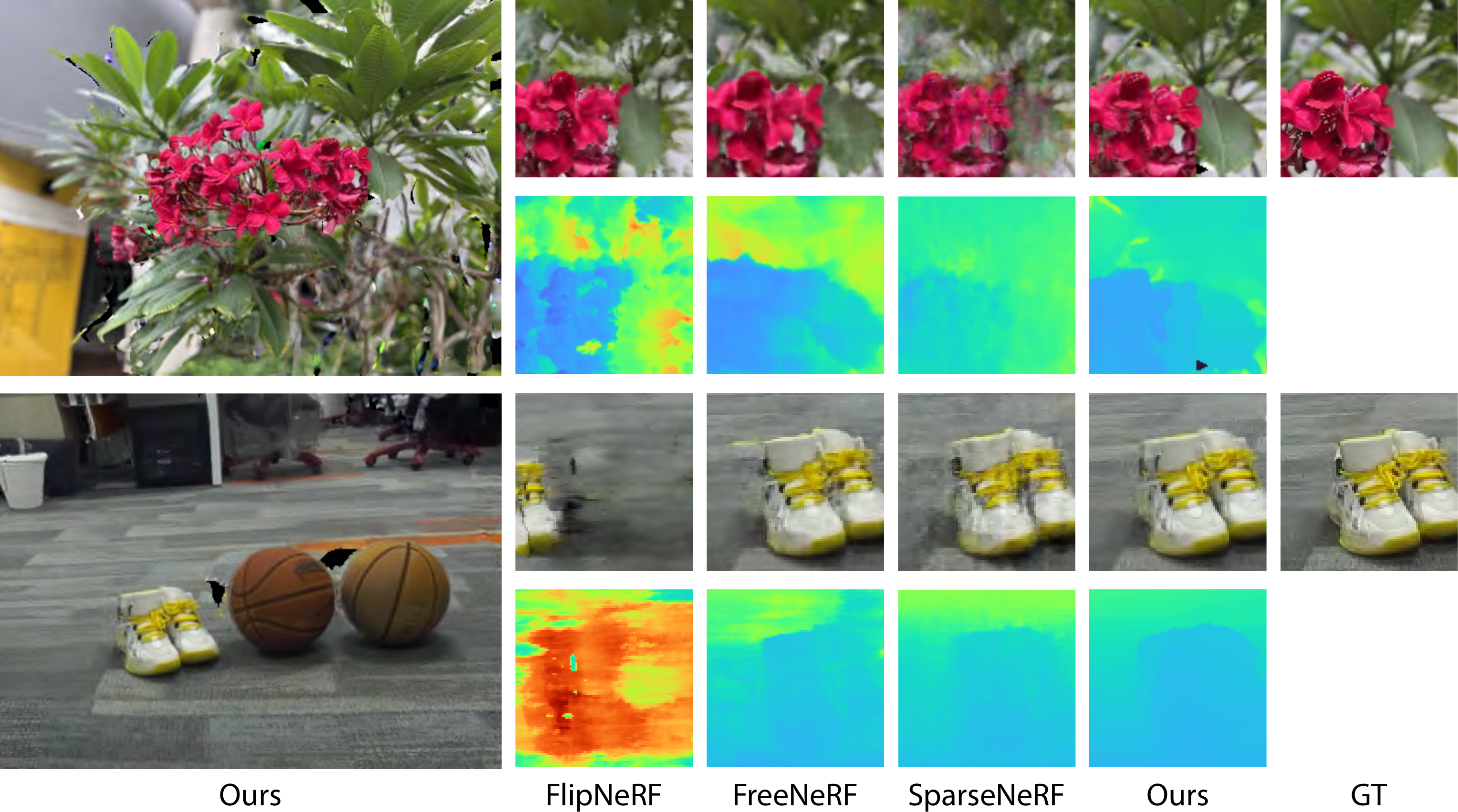}
  \vspace{-10pt}
  \caption{\textbf{NVS-RGBD 2 input (iPhone on the top and ZED 2 on the bottom)}. We show comparisons against other sparse-view NeRF-based approaches.}
  \label{fig:rgbd2}
  % \vspace{-10pt}
\end{figure}

\textbf{NVS-RGBD Dataset} \quad 
NVS-RGBD is a sparse input dataset with 2 and 3 views. We show the 2 input results in Fig.~\ref{fig:rgbd2}. On the Plant scene, the input views have a large angular difference in pose. NeRF-based approaches overfit to the training views and produce geometry and texture with significant artifacts. Our method is able to reconstruct a coherent geometry and high-quality texture in comparison. For the Basketball scene, SparseNeRF and FlipNeRF generate significant artifacts on the shoes while FreeNeRF is unable to capture texture details. Our approach produces details closer to ground truth without any noticeable artifacts. The differences are more prominent in the video comparisons provided in supplementary video.

\end{document}